\documentclass{article}
\usepackage{spconf,amsmath,epsfig}

\title{TOWARD TASK-DRIVEN SATELLITE IMAGE SUPER-RESOLUTION}
\name{\begin{tabular}{c}Maciej Ziaja$^{1,2}$, Pawel Kowaleczko$^{1,3}$, Daniel Kostrzewa$^{1,2}$ \\ {Nicolas Long\'{e}p\'{e}}$^{4}$, Michal Kawulok$^{1,2}$\end{tabular}
\thanks{This work was partially funded by European Space Agency. MK was supported by the National Science Centre, Poland, under Research Grant 2022/47/B/ST6/03009. MK was supported by the SUT funds through the Rector’s Research and Development Grant 02/080/RGJ24/0042.}}

\address{$^1$KP Labs, Gliwice, Poland\\$^2$Silesian University of Technology, Gliwice, Poland\\$^3$Warsaw University of Technology, Warsaw, Poland\\$^4$European Space Agency, Frascati, Italy\\\texttt{michal.kawulok@ieee.org}}

\begin{document}
%
\maketitle
\begin{abstract}
Super-resolution is aimed at reconstructing high-resolution images from low-resolution observations. State-of-the-art approaches underpinned with deep learning allow for obtaining outstanding results, generating images of high perceptual quality. However, it often remains unclear whether the reconstructed details are close to the actual ground-truth information and whether they constitute a more valuable source for image analysis algorithms. In the reported work, we address the latter problem, and we present our efforts toward learning super-resolution algorithms in a task-driven way to make them suitable for generating high-resolution images that can be exploited for automated image analysis. In the reported initial research, we propose a methodological approach for assessing the existing models that perform computer vision tasks in terms of whether they can be used for evaluating super-resolution reconstruction algorithms, as well as training them in a task-driven way. We support our analysis with experimental study and we expect it to establish a solid foundation for selecting appropriate computer vision tasks that will advance the capabilities of real-world super-resolution.
\end{abstract}
\section{Introduction}
\label{sec:intro}

Super-resolution (SR) reconstruction consists in processing a single~\cite{ChenHe2022} or multiple~\cite{Valsesia2022} images whose spatial resolution is insufficient, in order to generate a high-resolution (HR) image. As SR problem is ill-posed~\cite{Lugmayr2021space}, the goals of such operation are manifold and often contradictory~\cite{ChoiKim2020}---they can range from enhancing the perceptual image quality to discovering the actual HR details by integrating information present in the low-resolution (LR) input data. The former is aimed at obtaining images that appear as if they were real, but high-frequency details may be hallucinated rather than reconstructed---this is often the case for single-image SR (SISR), especially when large magnification ratios are applied, like $8\times$ or more. On the other hand, multi-image SR (MISR)~\cite{Nasrollahi2014} is based on a premise that each LR image carries a portion of HR information, which are integrated during the SR process. 

A common approach to evaluate the SR outcome is to compare it with an HR reference image---similarity between them is treated as the quality determinant. However, this is becoming challenging for real-world datasets~\cite{Benecki2018AA}, in which LR and HR images are captured at different spatial resolution, possibly originating from different satellites~\cite{Kowaleczko2023, Cornebise2022}. An interesting alternative is to use the super-resolved images as the input for further computer vision analysis and to compare the obtained results against those obtained after processing the original (or interpolated) LR image. However, this direction has been given relatively little attention---some attempts were reported for enhancing images of text documents~\cite{ChenLi2021} and also there were a few works reported in the field of remote sensing. In~\cite{Razzak2021}, Razzak et al. evaluated the HighRes-net MISR network~\cite{deudon2020highresnet} applied to super-resolve Sentinel-2 (S-2) images relying on building delineation. In our earlier work~\cite{Kawulok2023IGARSS}, we exploited several tasks aimed at retrieving physical parameters (such as nitrogen uptake or chlorophyll concentration) to verify the spectral consistency between the input images and the super-resolved ones. The task-based evaluation is particularly useful when an input LR image and the HR reference exhibit differences resulting from having been captured by different satellites. In such cases, the pixel-wise similarity metrics like peak signal-to-noise ratio (PSNR) or structural similarity index (SSIM) are not sufficiently robust for assessing the reconstruction accuracy~\cite{Kowaleczko2023}.

\begin{figure*}[!h]
    \centering
    \includegraphics[width=\textwidth]{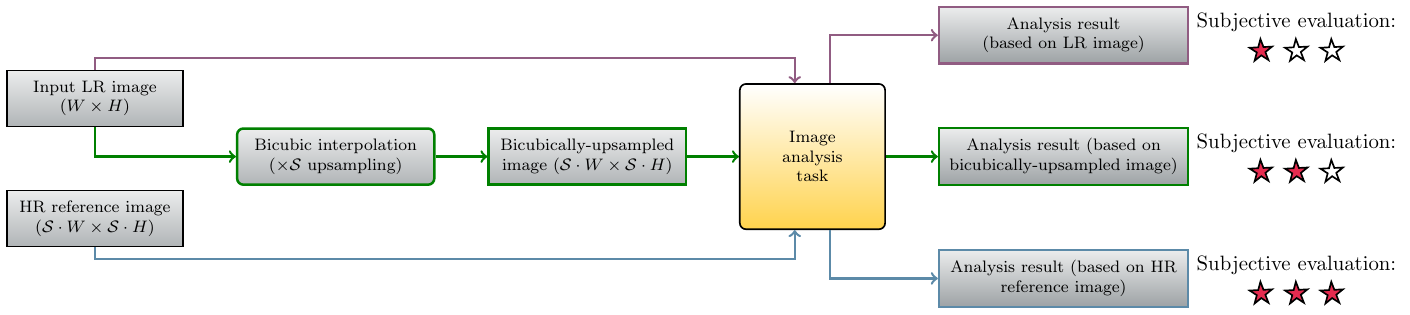}
    \caption{A general outline of the evaluation procedure. The image analysis is performed from the LR image (purple line), its bicubically-upsampled version (green line), and from the HR image (blue line). The obtained outcomes are assessed in a subjective manner, and the expected quality ranking is as shown on the right. The blocks with rounded corners indicate the actions, while the remaining ones---the generated artifacts.}
    \label{fig:flowchart}
\end{figure*}
Another possibility is to exploit the computer vision tasks during training a network that performs SR, so that the trained model is more useful in specific practical applications. Unfortunately, this interesting direction has not been explored intensively so far---some techniques were proposed for SISR of natural images, including  document scans, with object detection~\cite{Haris2021}, image segmentation~\cite{Frizza2022}, or optical character recognition~\cite{Madi2022} tasks used to guide the training process. However, to our best knowledge, such task-driven training procedures were not reported for super-resolving remotely-sensed images up to date. One of the main difficulties lies in selecting appropriate image analysis tasks and in preparing the ground-truth labels necessary to train the systems that perform the specific tasks. Also, the existing approaches to task-driven training were limited to using simulated datasets, in which LR images are obtained by degrading and downsampling HR references---as argued in~\cite{ChenHe2022}, such a scenario suffers from the domain performance gap: the models trained from simulated data are much less effective when presented real-world (i.e., not degraded) images. As argued later in this paper, elaborating appropriate models for task-driven training from real-world data is concerned with considerable challenges, including the lack of ground-truth annotations. 

Here, we report our initial study on preparing a set of diverse image processing tasks that will be exploited for training SISR and MISR networks in a task-driven manner for super-resolving S-2 images. We rely on the images from the MuS2 benchmark~\cite{Kowaleczko2023} which are coupled with HR references acquired during the WorldView-2 mission. Our contribution can be summarized as follows:
\begin{enumerate}\itemsep0em 
    \item We propose a systematic approach to assess the computer vision tasks in terms of their adequacy in evaluating SR outcome and guiding the training process. Importantly, we focus on exploiting the existing models that were trained beforehand without the necessity of elaborating ground-truth annotations for the scenes considered for SR.
    \item We report the results of our experimental study, in which we investigate four different image analysis techniques aimed at segmenting roads and building structures, detecting keypoints, and unsupervised image segmentation. 
    \item For two tasks related with roads and building segmentation, we show how to fine-tune the model to make it appropriate for the required scale. This includes adapting the available ground-truth target segmentation masks that we use to fine-tune the existing models. In this way, we avoid the necessity to prepare new annotations.
\end{enumerate}

\section{Methods}

An outline of our approach to assess the computer vision tasks in terms of their applicability to the SR domain is presented in Fig.~\ref{fig:flowchart}. We perform the task from three different inputs: (\textit{i})~the input LR image along with (\textit{ii})~its $\mathcal{S}\times$ bicubically-upsampled counterpart, and (\textit{iii})~the HR reference image. The obtained outcomes are evaluated on the qualitative basis---basically, the one obtained from the bicubically-upsampled image should be substantially better than that obtained from the LR image, but worse than that from the HR reference image. Such an order is critical, because the outcome obtained for the HR image is treated as the target in our setting---hence, the ground-truth annotations for the particular tasks are generated in an automatic manner, being in line with the self-supervised learning paradigm~\cite{LiuZhang2021}. 

For semantic image segmentation, we exploited a baseline U-Net architecture~\cite{Ronneberger2015} which was trained using the Massachusetts Buildings Dataset~\cite{Mnih2013} for building segmentation and using the Massachusetts Roads Dataset~\cite{Mnih2013} and DeepGlobe Road Extraction Dataset~\cite{Demir2018} for segmenting the roads. We have also considered the Key.Net network~\cite{Barroso2022} for keypoint detection and the Segment Anything (SA) network~\cite{Kirillov2023} for unsupervised segmentation. 
    
\begin{figure*}[!ht]
    \centering
    \renewcommand{\tabcolsep}{1mm}
    \begin{tabular}{ccc}
    \includegraphics[width=0.32\textwidth]{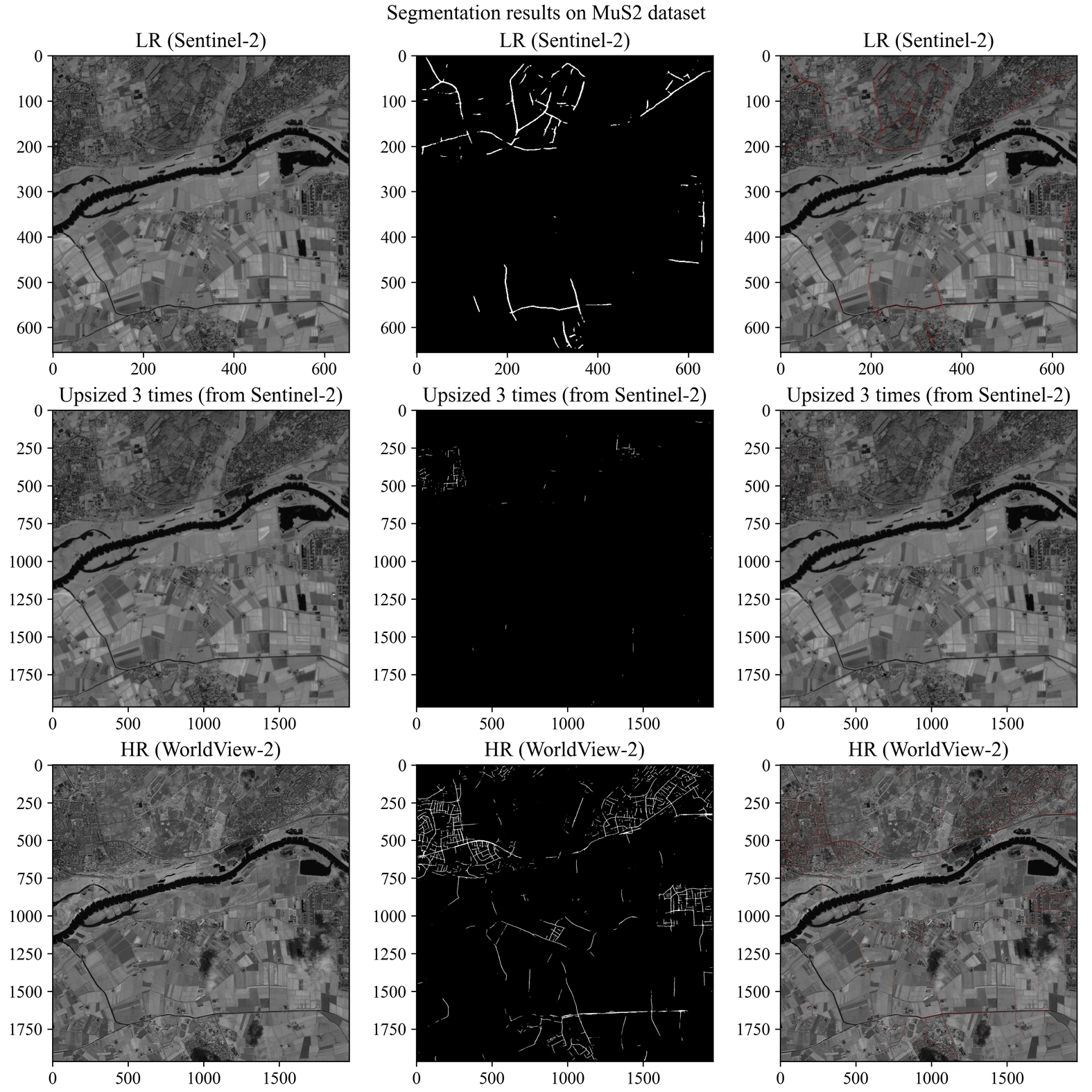}     &  
    \includegraphics[width=0.32\textwidth]{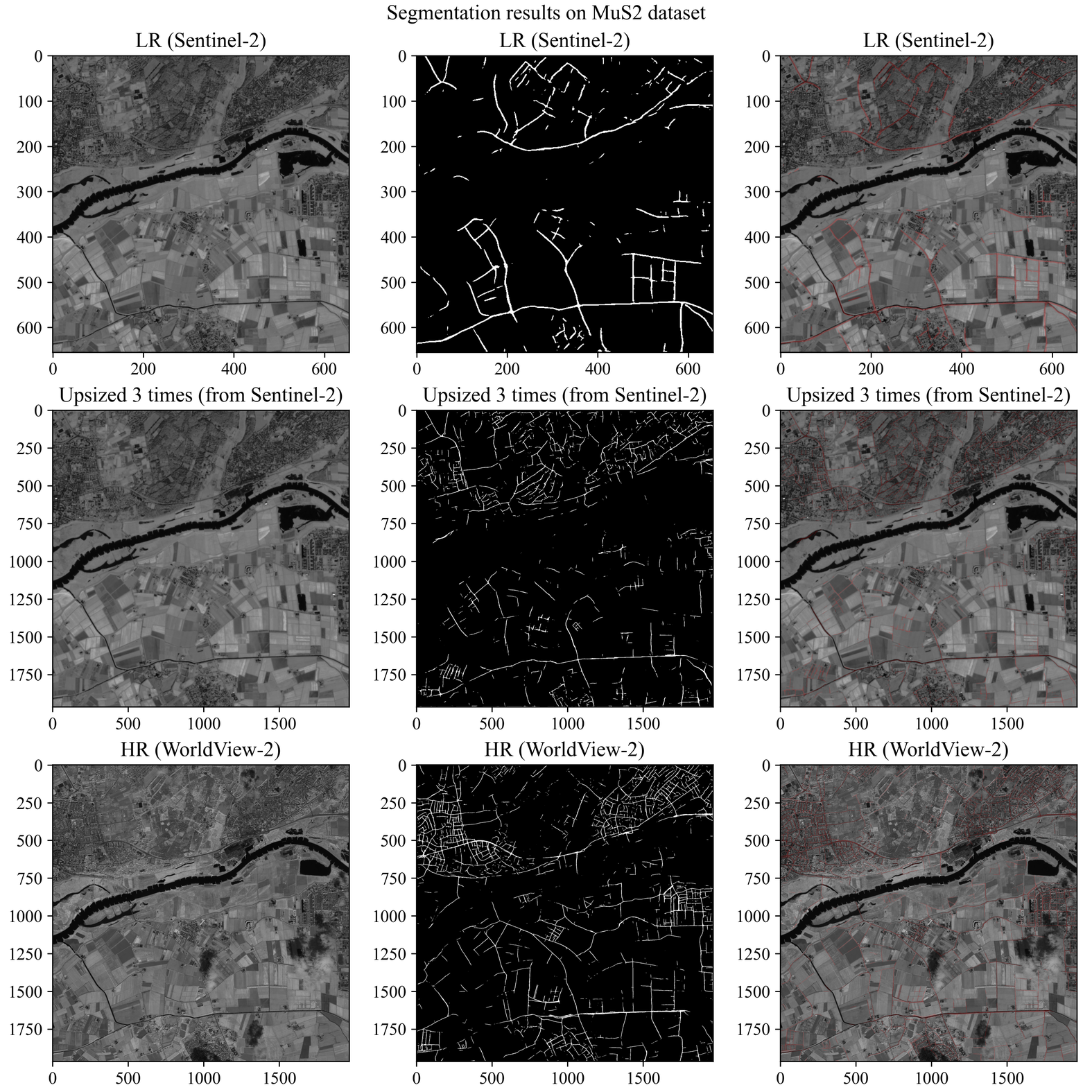}     &     
    \includegraphics[width=0.32\textwidth]{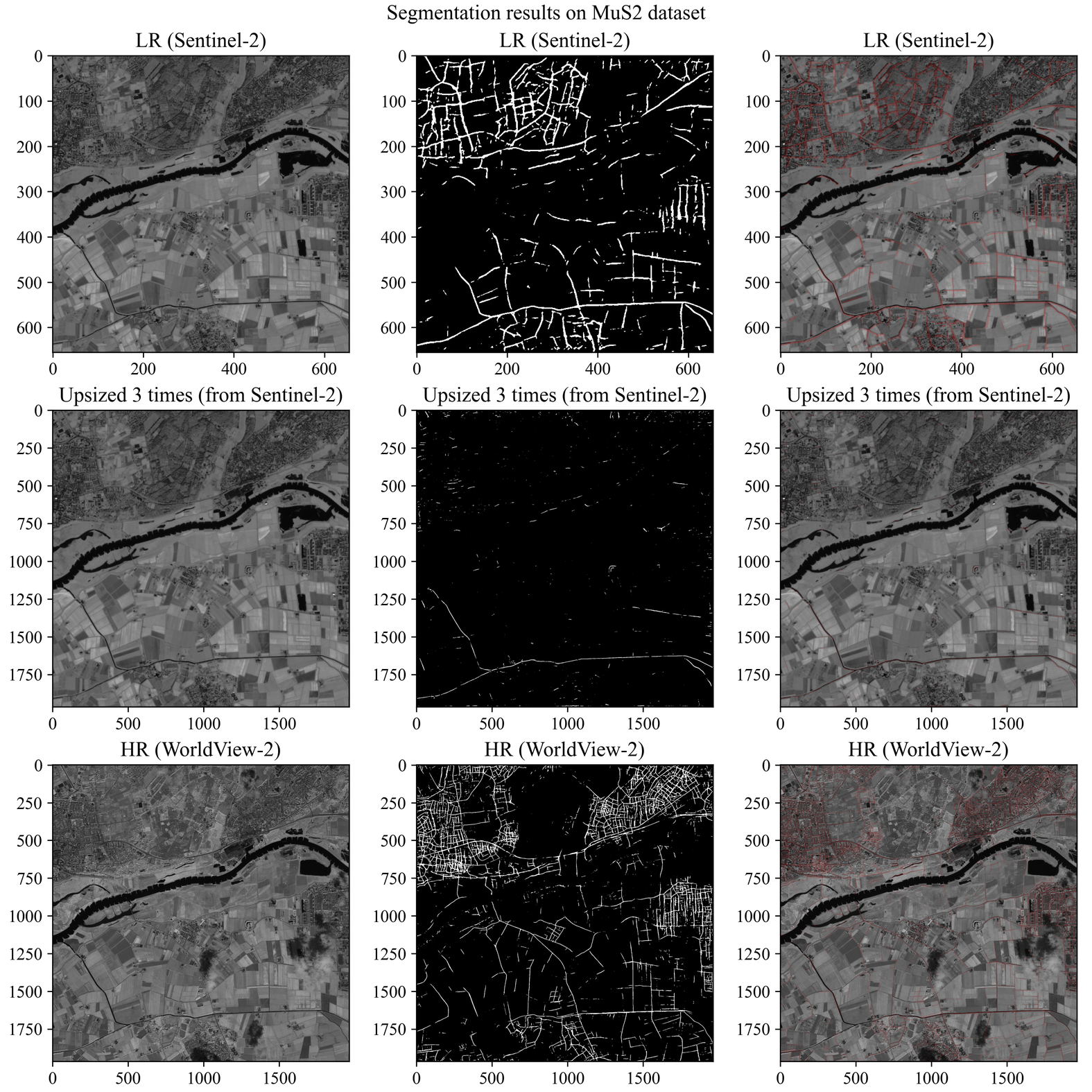}     \\
     (a) No adaptation & (b) Sample-wise adaptation & (c) Dataset-wise adaptation \\
    \end{tabular}
    
    \caption{Road segmentation performed for S-2 images (B08 band) from MuS2: original and bicubically-upsampled LR images, and the corresponding HR references, relying on different approaches to adapting the batch normalization parameters.}
    \label{fig:roads}
\end{figure*}
    

While for Key.Net and SA, we exploit the already-trained models~\cite{Barroso2022,Kirillov2023}, the experiments (Section~\ref{sec:exp}) revealed that the U-Nets are highly scale-sensitive and if they are trained from the images at 1\,m ground sampling distance (GSD), then they are not suitable for segmenting S-2 images before (10\,m GSD) and after applying SR (3.3\,m GSD). As our goal was to exploit the existing models (and the original annotated training data), we downsampled the images from the aforementioned datasets~\cite{Demir2018,Mnih2013} along with the ground-truth annotations to match the scale of 3.3\,m GSD, and we retrained the U-Net models from the adjusted data. Moreover, as the images used for training have different spectral properties than the images in the MuS2 dataset, we found it necessary to adjust the parameters in the U-Net batch normalization layers. We considered two variants: (\textit{i})~\textit{sample-wise adaptation}, in which the parameters are adjusted for every image that is processed, and (\textit{ii})~\textit{dataset-wise adaptation}, in which the whole dataset is processed to determine the batch normalization parameters. Also, we augmented the training data by applying intensity inversion to randomly selected samples---this improved the models' generalization. 

\section{Experiments} \label{sec:exp}

In our experiments, we used the images from the MuS2 dataset~\cite{Kowaleczko2023} which contains 91 scenes, each of which is composed of 14--15 S-2 images showing the same region of interest, coupled with a WorldView-2 image downsampled to a size that is $\mathcal{S}=3\times$ larger than the 10\,m GSD S-2 bands. These images are supposed to be later used for task-driven training. For training the road and building segmentation, we exploited the Massachusetts Dataset~\cite{Mnih2013} with ground-truth masks showing buildings and roads, and DeepGlobe Road Extraction Dataset~\cite{Demir2018}.

\begin{figure*}[!h]
    \centering
    \renewcommand{\tabcolsep}{1mm}
    \begin{tabular}{ccc}
    \includegraphics[width=0.32\textwidth]{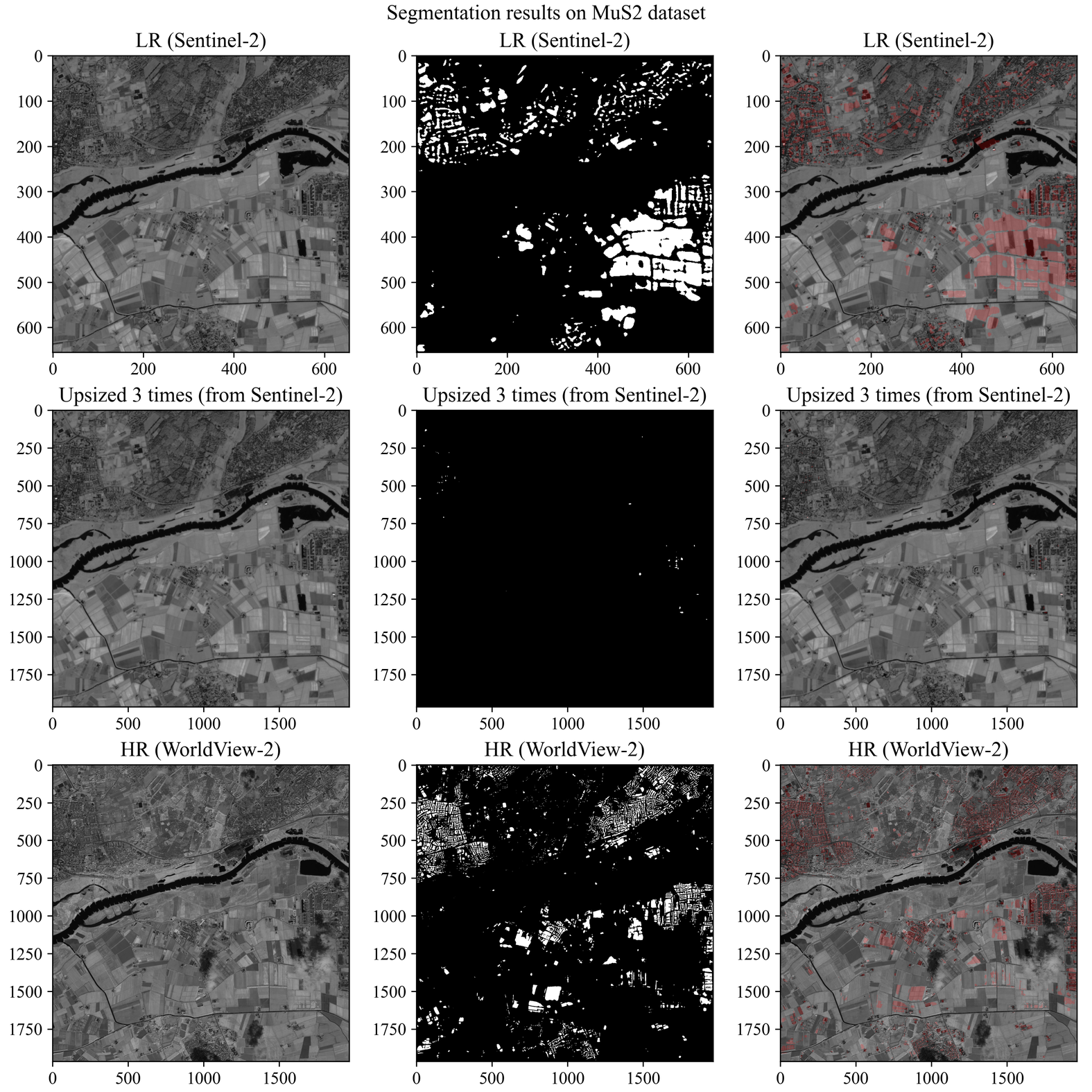}     &  
    \includegraphics[width=0.32\textwidth]{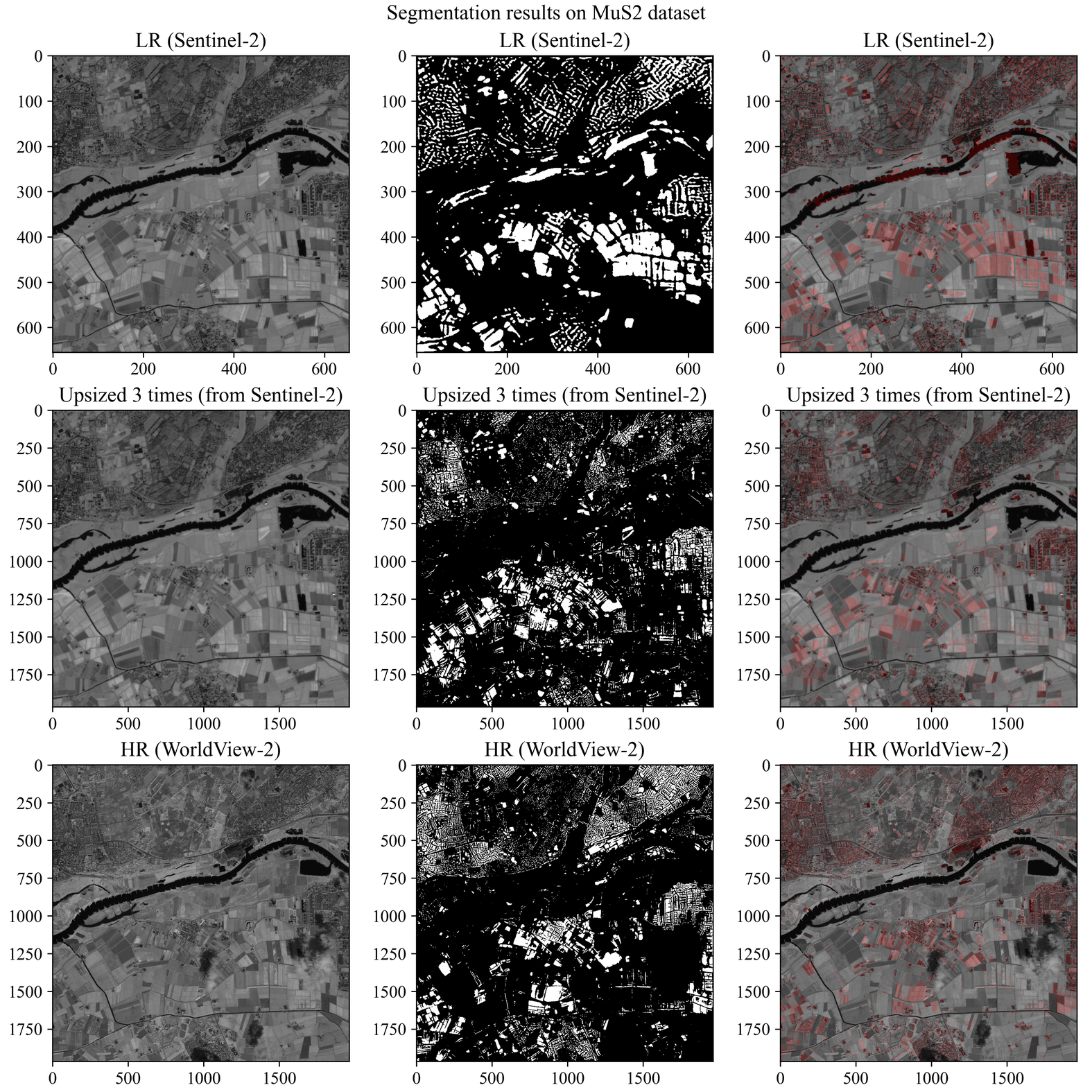}     &      
    \includegraphics[width=0.32\textwidth]{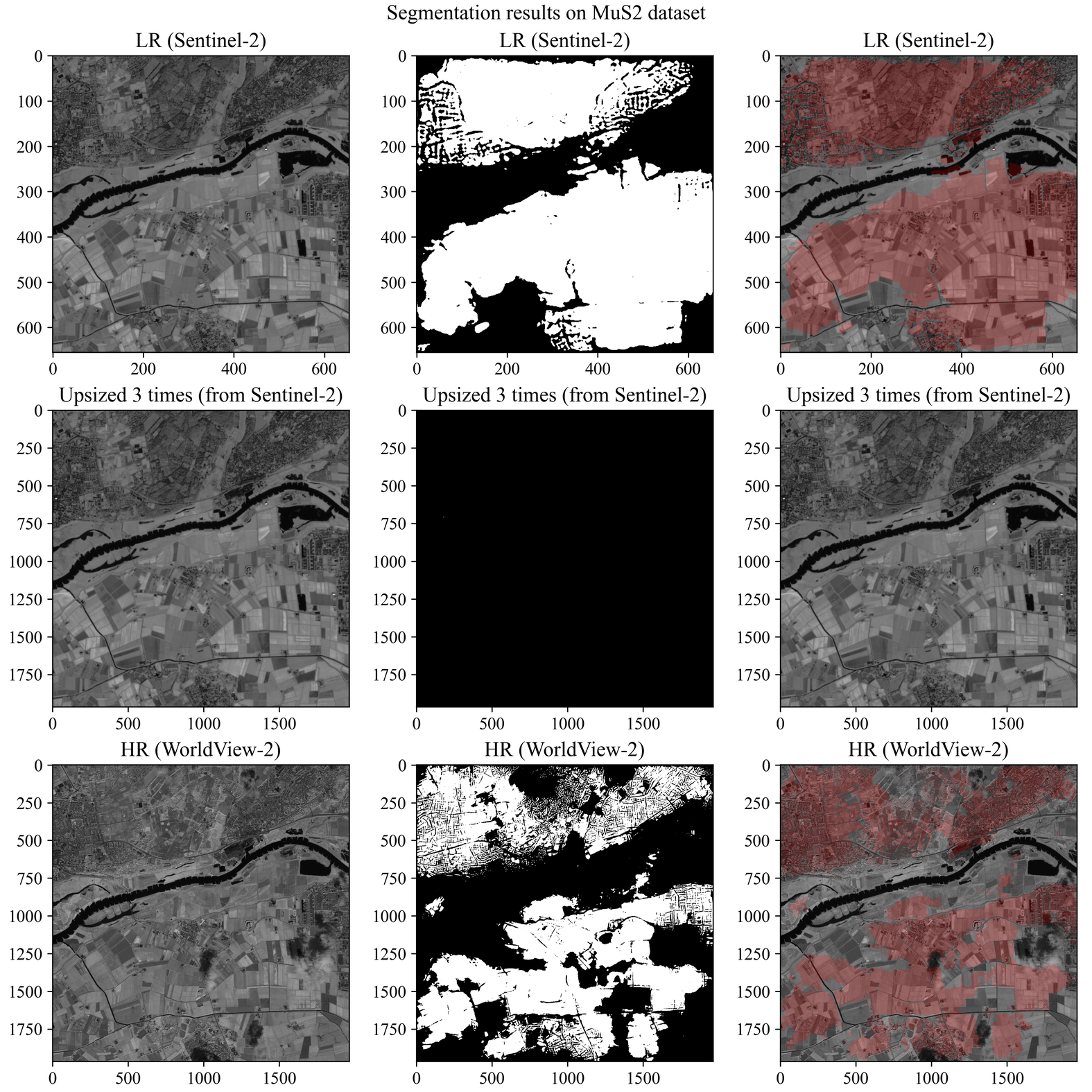}     \\
     (a) No adaptation & (b) Sample-wise adaptation & (c) Dataset-wise adaptation \\
    \end{tabular}
    
    \caption{Building segmentation performed for S-2 images (B08 band) from the MuS2 benchmark: original and bicubically-upsampled LR images, and the corresponding HR references, relying on different approaches to adapting the batch normalization parameters.}
    \label{fig:buildings}
\end{figure*}

In Fig.~\ref{fig:roads}, we present examples of road segmentation performed with a U-Net trained without any adaptation, followed with sample-wise (b) and dataset-wise adaptation (c) of batch normalization parameters. For dataset-wise adaptation, the road network is visually better captured than for sample-wise adaptation, but the outcome obtained for the bicubically-upsampled image is worse than for that retrieved from the LR image. In the case of sample-wise adaptation, it can be concluded that the order is appropriate. It can also be seen that the results are rather poor without adapting the batch normalization parameters~(a), despite standardization of the input data. Importantly, the presented example is representative for the whole MuS2 dataset---although our approach does not include any quantitative analysis, the qualitative assessment is sufficiently clear in this case. In Fig.~\ref{fig:buildings}, we present an analogous example for segmenting building structures. Here, the dataset-wise adaptation fails completely, and the sample-wise adaptation provides satisfactory results (consistently for the whole dataset). Importantly, the models trained from the images at their original GSD were not capable of detecting any roads or buildings, but after adjusting the scale of the images in the training set, it is possible to achieve the required result. If the SR network is trained in a task-driven way so that the masks of the road network and building structures extracted from the super-resolved images are similar to the masks retrieved from the HR images, we can expect that other details in the super-resolved images will get closer to the actual ground-truth. Furthermore, such images should constitute a valuable input also for other image analysis tasks that rely on high-frequency features.

\begin{figure}[t]
    \centering
    \renewcommand{\tabcolsep}{1mm}
    \resizebox{\columnwidth}{!}{
    \begin{tabular}{ccc}
        \includegraphics[width=0.31\columnwidth]{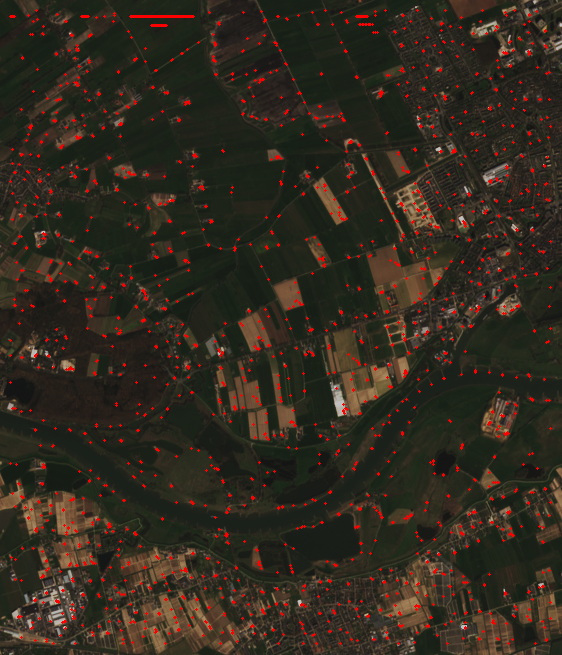} & 
        \includegraphics[width=0.31\columnwidth]{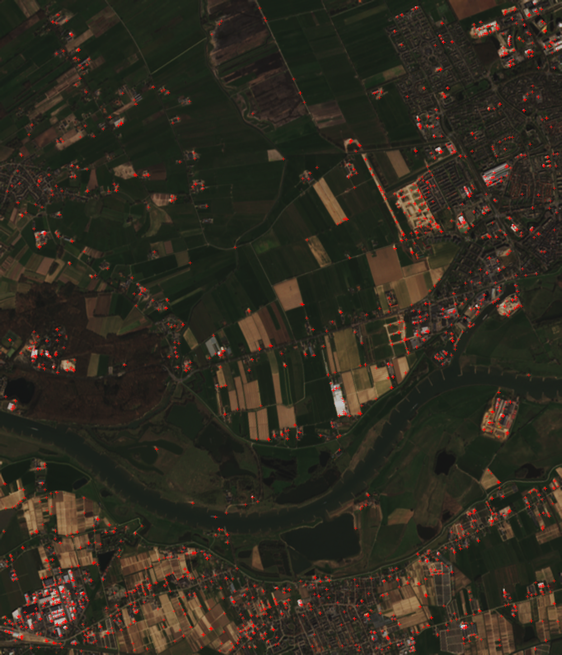} & 
        \includegraphics[width=0.31\columnwidth]{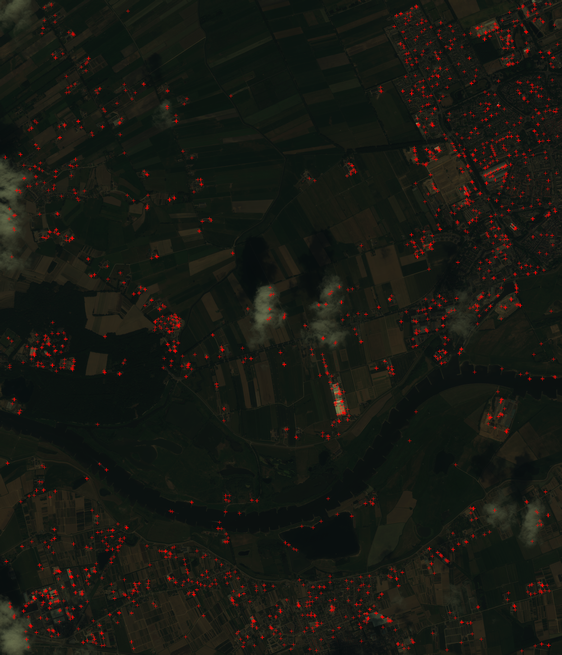} \\
        (a)  & (b)  & (c)  \\
    \end{tabular}
    }
    \caption{Keypoints retrieved with Key.Net for (a)~original S-2 image, (b)~bicubically-upsampled image (by a factor of $3\times$, and (c)~WorldView-2 HR reference. The color images were composed from S-2 B02, B03 and B04 bands at 10\,m GSD.}
    \label{fig:keynet}
\end{figure}

\begin{figure}[t]
    \centering
    \renewcommand{\tabcolsep}{1mm}
    \resizebox{\columnwidth}{!}{
    \begin{tabular}{cccc}
        \includegraphics[width=0.23\columnwidth]{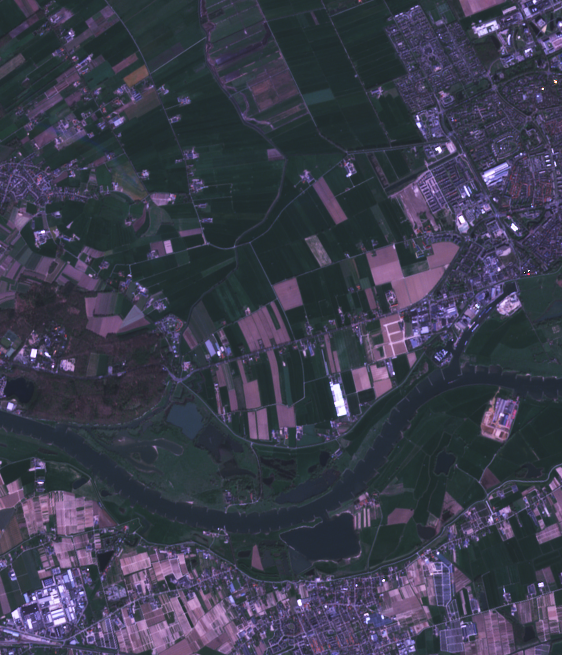} & 
        \includegraphics[width=0.23\columnwidth]{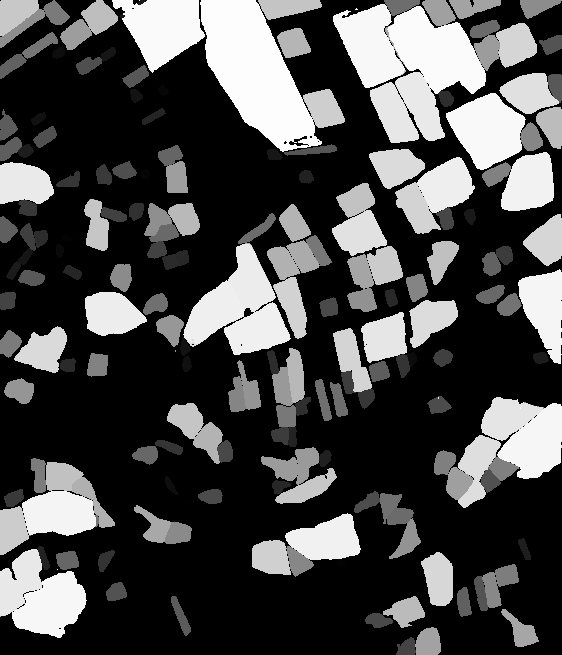} & 
        \includegraphics[width=0.23\columnwidth]{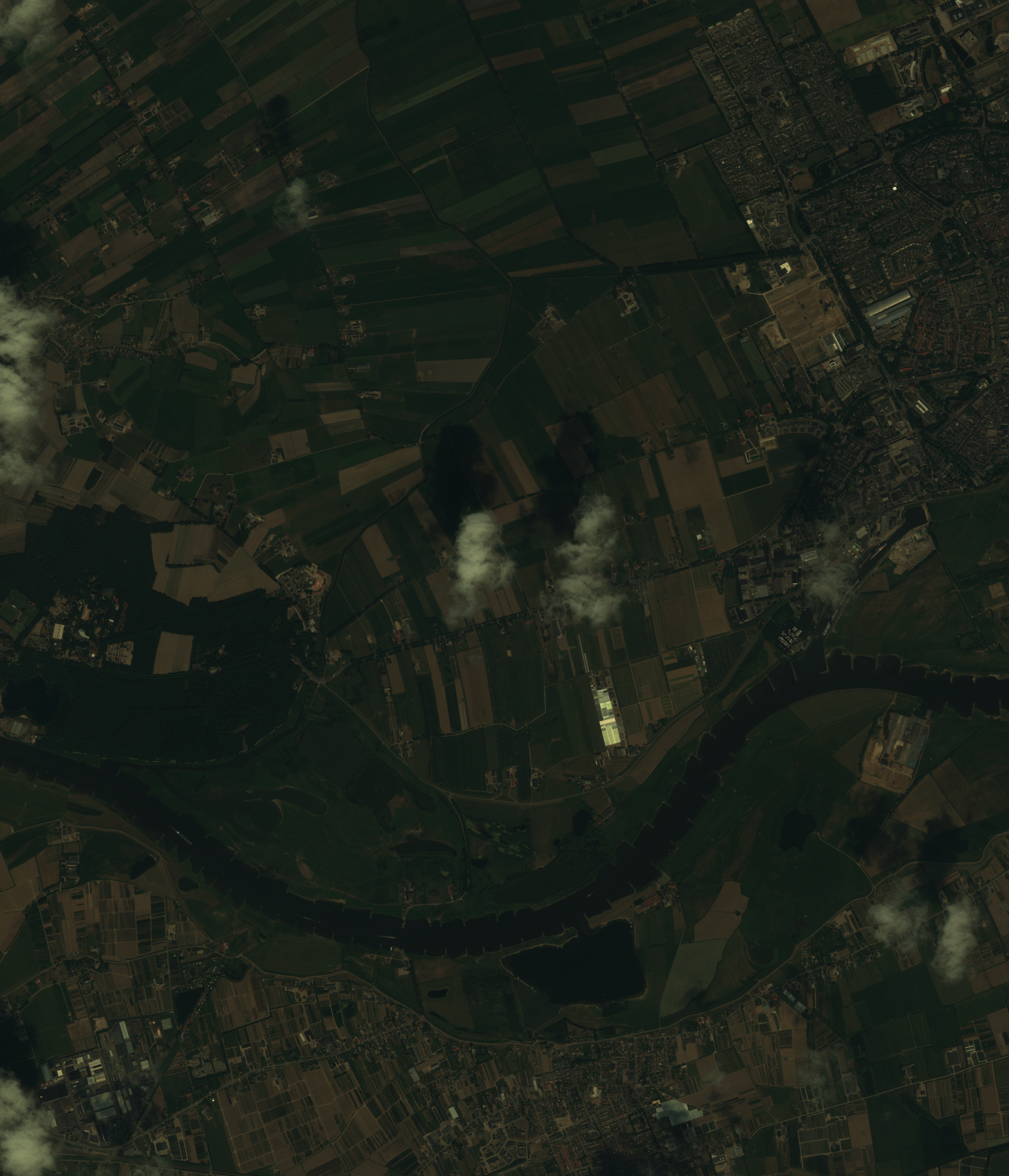} &
        \includegraphics[width=0.23\columnwidth]{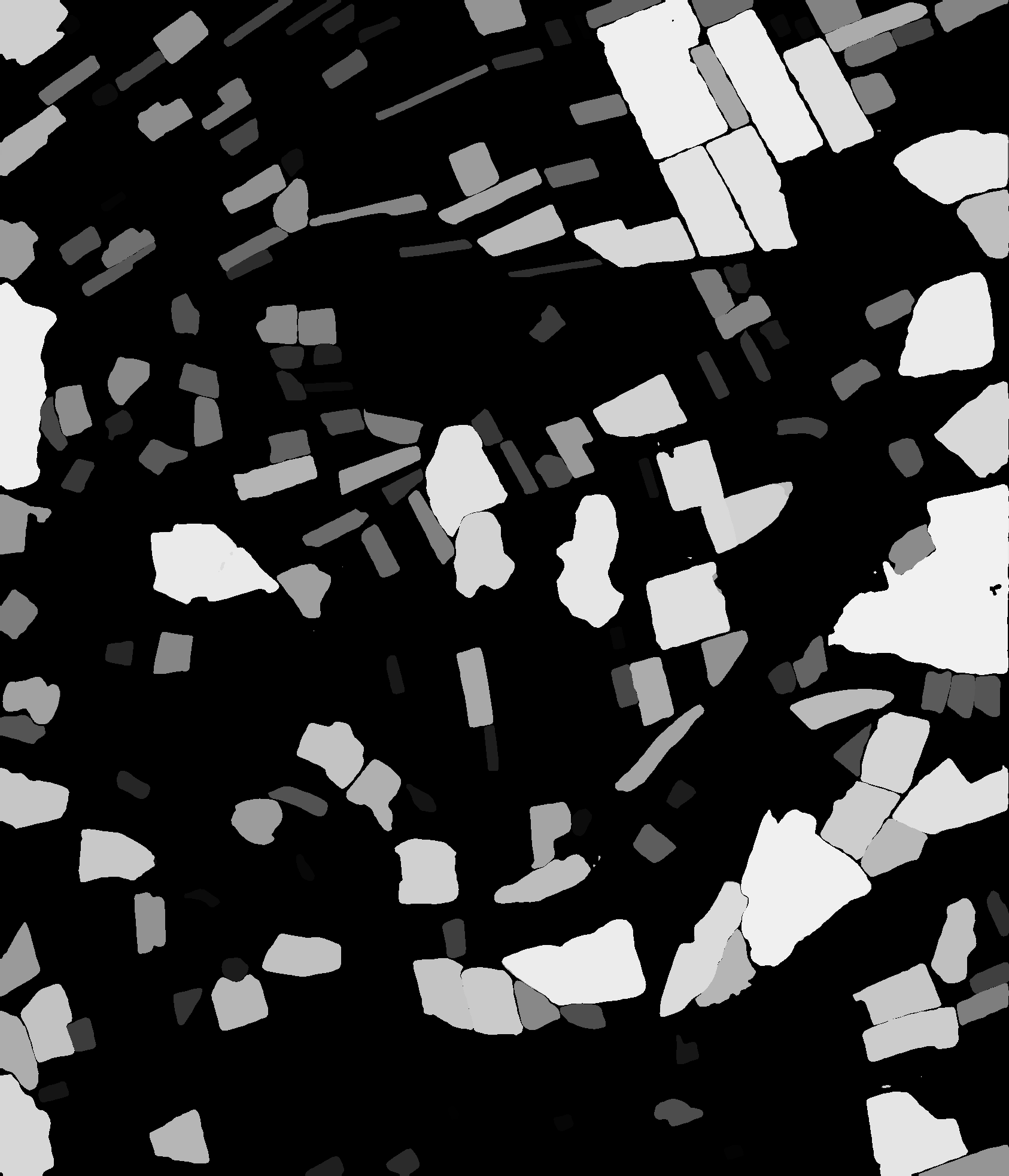} \\
        (a) & (b) & (c) & (d) \\
    \end{tabular}
    }
    \caption{Results of unsupervised segmentation obtained with SA for an S-2 image~(a, b) and for the HR reference~(c, d).}
    \label{fig:sam}
\end{figure}

In Fig.~\ref{fig:keynet}, we show the keypoints detected using the original Key.Net model~\cite{Barroso2022} (for every images, the same number of 1000 points is localized). It can be appreciated that while in the LR image~(a) the feature points are evenly spaced, in the bicubically-upsampled image~(b) and in the HR reference~(c), the points are located in the urban regions characterized with more high-frequency details. In our opinion, the high-frequency details are more accurately captured in the HR image, so driving the detection result from~(b) to~(c) should positively influence the reconstruction quality. Contrary to that, the results obtained with unsupervised segmentation performed using SA (Fig.~\ref{fig:sam}) do not lead to unambiguous conclusions---the results for LR S-2 images and HR WorldView-2 reference manifest some similarities, but apparently the segmentation relies on low-frequency features. Possibly exploiting such a task may be helpful in regularizing the SR training (by controlling the consistency of smooth areas in the images), but perceptually the outcome for HR images is not ``better'' than that for the LR image.


\section{Conclusions and outlook}

In this paper, we reported our preliminary study to elaborate a battery of image analysis tasks that could be used for evaluating and training deep networks for SR. We demonstrated that the existing models can be adapted to a specific scale without the need for creating ground-truth annotations. Moreover, by adapting the batch normalization parameters, we can effectively deal with spectral differences between the images used for training the task-specific networks  (e.g., U-Net for segmenting the roads) and the images used for training the SR networks. However, the choice of the granularity for such adaptation should be done individually for each task.

Our ongoing work is focused on exploring how to exploit these tasks as loss functions for training deep networks that realize MISR of S-2 images~\cite{Tarasiewicz2023TGRS}. The investigated options include comparing the final outcome in the same manner as when training a task-specific network, but we also consider defining the loss functions in a feature space---instead of comparing the final outcomes, we can maximize the similarity between the features that are extracted from two images in order to perform the task. Furthermore, our initial attempts suggest that it is beneficial to combine the task-driven loss functions with the image-based ones (L1 or L2) that are commonly used for training SR networks---this increases the convexity and helps avoid getting stuck in local minima. Such image-based loss functions can be computed between the super-resolved image and the HR reference, but also between the input LR image and the downsampled SR outcome to ensure the consistency. The latter is particularly useful for real-world datasets, where HR images differ from the LR images not only in terms of spatial resolution, but also due to different sensor characteristics.



\begin{thebibliography}{10}
\providecommand{\url}[1]{#1}
\csname url@samestyle\endcsname
\providecommand{\newblock}{\relax}
\providecommand{\bibinfo}[2]{#2}
\providecommand{\BIBentrySTDinterwordspacing}{\spaceskip=0pt\relax}
\providecommand{\BIBentryALTinterwordstretchfactor}{4}
\providecommand{\BIBentryALTinterwordspacing}{\spaceskip=\fontdimen2\font plus
\BIBentryALTinterwordstretchfactor\fontdimen3\font minus \fontdimen4\font\relax}
\providecommand{\BIBforeignlanguage}[2]{{%
\expandafter\ifx\csname l@#1\endcsname\relax
\typeout{** WARNING: IEEEtranS.bst: No hyphenation pattern has been}%
\typeout{** loaded for the language `#1'. Using the pattern for}%
\typeout{** the default language instead.}%
\else
\language=\csname l@#1\endcsname
\fi
#2}}
\providecommand{\BIBdecl}{\relax}
\BIBdecl

\bibitem{Barroso2022}
A.~Barroso-Laguna and K.~Mikolajczyk, ``{Key.Net}: Keypoint detection by handcrafted and learned {CNN} filters revisited,'' \emph{IEEE Transactions on Pattern Analysis and Machine Intelligence}, vol.~45, no.~1, pp. 698--711, 2022.

\bibitem{Benecki2018AA}
P.~Benecki, M.~Kawulok, D.~Kostrzewa, and L.~Skonieczny, ``Evaluating super-resolution reconstruction of satellite images,'' \emph{Acta Astronautica}, vol. 153, pp. 15--25, 2018.

\bibitem{ChenHe2022}
H.~Chen, X.~He, L.~Qing, Y.~Wu, C.~Ren, R.~E. Sheriff, and C.~Zhu, ``Real-world single image super-resolution: A brief review,'' \emph{Information Fusion}, vol.~79, pp. 124--145, 2022.

\bibitem{ChenLi2021}
J.~Chen, B.~Li, and X.~Xue, ``Scene text telescope: Text-focused scene image super-resolution,'' in \emph{Proc. IEEE/CVF CVPR}, 2021, pp. 12\,026--12\,035.

\bibitem{ChoiKim2020}
J.-H. Choi, J.-H. Kim, M.~Cheon, and J.-S. Lee, ``Deep learning-based image super-resolution considering quantitative and perceptual quality,'' \emph{Neurocomputing}, vol. 398, pp. 347--359, 2020.

\bibitem{Cornebise2022}
J.~Cornebise, I.~Orsolic, and F.~Kalaitzis, ``Open high-resolution satellite imagery: The {WorldStrat} dataset {\textendash} with application to super-resolution,'' in \emph{Proc. NeurIPS}, 2022.

\bibitem{Demir2018}
I.~Demir, K.~Koperski, D.~Lindenbaum, G.~Pang, J.~Huang, S.~Basu, F.~Hughes, D.~Tuia, and R.~Raskar, ``Deepglobe 2018: A challenge to parse the {Earth} through satellite images,'' in \emph{Proc. IEEE/CVF CVPR Workshops}, 2018, pp. 172--181.

\bibitem{deudon2020highresnet}
M.~Deudon, A.~Kalaitzis, I.~Goytom, M.~R. Arefin, Z.~Lin, K.~Sankaran, V.~Michalski, S.~E. Kahou, J.~Cornebise, and Y.~Bengio, ``{HighRes}-net: Recursive fusion for multi-frame super-resolution of satellite imagery,'' \emph{arXiv preprint arXiv:2002.06460}, 2020.

\bibitem{Frizza2022}
T.~Frizza, D.~G. Dansereau, N.~M. Seresht, and M.~Bewley, ``Semantically accurate super-resolution generative adversarial networks,'' \emph{Computer Vision and Image Understanding}, p. 103464, 2022.

\bibitem{Haris2021}
M.~Haris, G.~Shakhnarovich, and N.~Ukita, ``Task-driven super resolution: Object detection in low-resolution images,'' in \emph{Proc. ICONIP}.\hskip 1em plus 0.5em minus 0.4em\relax Springer, 2021, pp. 387--395.

\bibitem{Kawulok2023IGARSS}
M.~Kawulok, P.~Kowaleczko, M.~Ziaja, J.~Nalepa, D.~Kostrzewa, D.~Latini, D.~De~Santis, G.~Salvucci, I.~Petracca, V.~La~Pegna, Z.~Bartalis, and F.~Del~Frate, ``Understanding the value of hyperspectral image super-resolution from {PRISMA} data,'' in \emph{Proc. IEEE IGARSS}, 2023, pp. 1489--1492.

\bibitem{Kirillov2023}
A.~Kirillov, E.~Mintun, N.~Ravi, H.~Mao, C.~Rolland, L.~Gustafson, T.~Xiao, S.~Whitehead, A.~C. Berg, W.-Y. Lo \emph{et~al.}, ``Segment anything,'' \emph{arXiv preprint arXiv:2304.02643}, 2023.

\bibitem{Kowaleczko2023}
P.~Kowaleczko, T.~Tarasiewicz, M.~Ziaja, D.~Kostrzewa, J.~Nalepa, P.~Rokita, and M.~Kawulok, ``A real-world benchmark for {Sentinel-2} multi-image super-resolution,'' \emph{Scientific Data}, vol.~10, no.~1, p. 644, 2023.

\bibitem{LiuZhang2021}
X.~Liu, F.~Zhang, Z.~Hou, L.~Mian, Z.~Wang, J.~Zhang, and J.~Tang, ``Self-supervised learning: Generative or contrastive,'' \emph{IEEE Transactions on Knowledge and Data Engineering}, 2021.

\bibitem{Lugmayr2021space}
A.~Lugmayr, M.~Danelljan, and R.~Timofte, ``{NTIRE} 2021 learning the super-resolution space challenge,'' in \emph{Proc. IEEE/CVF CVPR}, 2021, pp. 596--612.

\bibitem{Madi2022}
B.~Madi, R.~Alaasam, and J.~El-Sana, ``Text edges guided network for historical document super resolution,'' in \emph{Proc. Int. Conf. on Frontiers in Handwriting Recognition}.\hskip 1em plus 0.5em minus 0.4em\relax Springer, 2022, pp. 18--33.

\bibitem{Mnih2013}
V.~Mnih, \emph{Machine learning for aerial image labeling}.\hskip 1em plus 0.5em minus 0.4em\relax University of Toronto (Canada), 2013.

\bibitem{Nasrollahi2014}
K.~Nasrollahi and T.~B. Moeslund, ``Super-resolution: A comprehensive survey,'' \emph{Machine Vision and Applications}, vol.~25, no.~6, pp. 1423--1468, 2014.

\bibitem{Razzak2021}
M.~T. Razzak, G.~Mateo-Garc{\'\i}a, G.~Lecuyer, L.~G{\'o}mez-Chova, Y.~Gal, and F.~Kalaitzis, ``Multi-spectral multi-image super-resolution of {Sentinel-2} with radiometric consistency losses and its effect on building delineation,'' \emph{ISPRS Journal of Photogrammetry and Remote Sensing}, vol. 195, pp. 1--13, 2023.

\bibitem{Ronneberger2015}
O.~Ronneberger, P.~Fischer, and T.~Brox, ``{U-Net}: Convolutional networks for biomedical image segmentation,'' in \emph{Medical Image Computing and Computer-Assisted Intervention--MICCAI 2015: 18th International Conference, Munich, Germany, October 5-9, 2015, Proceedings, Part III 18}.\hskip 1em plus 0.5em minus 0.4em\relax Springer, 2015, pp. 234--241.

\bibitem{Tarasiewicz2023TGRS}
T.~Tarasiewicz, J.~Nalepa, R.~A. Farrugia, G.~Valentino, M.~Chen, J.~A. Briffa, and M.~Kawulok, ``Multitemporal and multispectral data fusion for super-resolution of {Sentinel-2} images,'' \emph{IEEE Transactions on Geoscience and Remote Sensing}, vol.~61, pp. 1--19, 2023.

\bibitem{Valsesia2022}
D.~Valsesia and E.~Magli, ``Permutation invariance and uncertainty in multitemporal image super-resolution,'' \emph{IEEE Transactions on Geoscience and Remote Sensing}, vol.~60, pp. 1--12, 2022.

\end{thebibliography}


\end{document}